\documentclass[biblatex,english]{lni} 
\usepackage{booktabs}

\usepackage[]{blindtext}

\usepackage[nolist]{acronym} 
\usepackage{subcaption} 
\newcommand{\RNum}[1]{\uppercase\expandafter{\romannumeral #1\relax}} 
\usepackage{wrapfig} 
\usepackage{float} 

\usepackage{tabularx}  
\usepackage{setspace} 
\newcolumntype{Y}{>{\RaggedRight\arraybackslash}X} 
\usepackage{ragged2e}  
\usepackage{enumitem} 

\usepackage{tikz} 
\newcommand{\dittotikz}{%
    \tikz{
        \draw [line width=0.12ex] (-0.2ex,0) -- +(0,0.8ex)
            (0.2ex,0) -- +(0,0.8ex);
        \draw [line width=0.08ex] (-0.6ex,0.4ex) -- +(-1em,0)
            (0.6ex,0.4ex) -- +(1em,0);
    }%
}

\begin{document}
\title{Assessing the Impact of Image Dataset Features on Privacy-Preserving Machine Learning}
\author[1]{Lucas Lange}{lange@informatik.uni-leipzig.de}{0000-0002-6745-0845}
\author[1]{Maurice-Maximilian Heykeroth}{mh40qyqu@studserv.uni-leipzig.de}{}
\author[1]{Erhard Rahm}{rahm@informatik.uni-leipzig.de}{0000-0002-2665-1114}
\affil[1]{Leipzig University \& ScaDS.AI Dresden/Leipzig, Germany}
\maketitle

\begin{abstract}
Machine Learning (ML) is crucial in many sectors, including computer vision. However, ML models trained on sensitive data face security challenges, as they can be attacked and leak information. Privacy-Preserving Machine Learning (PPML) addresses this by using Differential Privacy (DP) to balance utility and privacy. This study identifies image dataset characteristics that affect the utility and vulnerability of private and non-private Convolutional Neural Network (CNN) models. Through analyzing multiple datasets and privacy budgets, we find that imbalanced datasets increase vulnerability in minority classes, but DP mitigates this issue. Datasets with fewer classes improve both model utility and privacy, while high entropy or low Fisher Discriminant Ratio (FDR) datasets deteriorate the utility-privacy trade-off. These insights offer valuable guidance for practitioners and researchers in estimating and optimizing the utility-privacy trade-off in image datasets, helping to inform data and privacy modifications for better outcomes based on dataset characteristics.
\end{abstract}


\begin{acronym}
    \setlength{\parskip}{0ex}
    \setlength{\itemsep}{-7px}
    \acro{ML}{Machine Learning}
    \acro{AI}{Artificial Intelligence}
    \acro{MLaaS}{Machine Learning as a Service}
    \acro{GPU}{Graphical Processing Unit}
    \acro{PPML}{Privacy-Preserving Machine Learning}
    \acro{DP}{Differential Privacy}
    \acro{MIA}{Membership Inference Attack}
    \acro{LiRA}{Likelihood Ratio Attack}
    \acro{CNN}{Convolutional Neural Network}
    \acro{SGD}{Stochastic Gradient Descent}
    \acro{SGDM}{Stochastic Gradient Descent with Momentum}
    \acro{ReLU}{Rectified Linear Unit}
    \acro{MNIST}{Modified National Institute of Standards and Technology}
    \acro{FMNIST}{Fashion MNIST}
    \acro{CIFAR}{Canadian Institute For Advanced Research}
    \acro{SVHN}{Street View House Numbers}
    \acro{EMNIST}{Extended MNIST}
    \acro{AUC}{Area Under Curve}
    \acro{FPR}{False Positive Rate}
    \acro{TPR}{True Positive Rate}
    \acro{ROC}{Receiver Operator Characteristics}
    \acro{FDR}{Fisher Discriminant Ratio}
    \acro{STD}{Standard Deviation}
    \acro{TP}{True Positives}
    \acro{TN}{True Negatives}
    \acro{FP}{False Positives}
    \acro{FN}{False Negatives}
    \acro{JPEG}{Joint Photographic Experts Group}
    \acro{PNG}{Portable Network Graphics}
\end{acronym}

\section{Introduction}
\label{sec:introduction}

\begin{figure}[t]
    \centering
    \begin{subfigure}[t]{0.27\textwidth}
    \includegraphics[width=\textwidth]{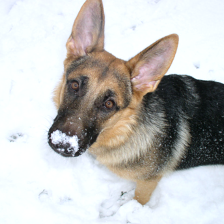}
    \caption{Original training image.}
    \end{subfigure}
    \hspace{2cm}
    \begin{subfigure}[t]{0.27\textwidth}
    \includegraphics[width=\textwidth]{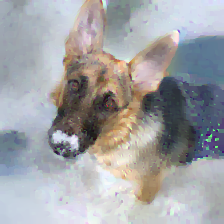}
    \caption{Reconstructed image.}
    \end{subfigure}
    \caption[Result of a Model Inversion Attack Using Gradient Information]{
    Result of a model inversion attack in a federated learning scenario using gradient information.
    Left image shows the original image that was used to train the model, the right image shows the reconstructed image from an inversion attack. Results of the reconstruction attack by~\cite{Geiping_Bauermeister_Dröge_Moeller_2020}.}
    \label{fig:inverting_gradient_information_attack}
    \end{figure}

Since the success of applications like ChatGPT\footnote{\url{https://openai.com/index/chatgpt/}} and DALL-E 2\footnote{\url{https://openai.com/index/dall-e-3/}}, new \ac{AI} technologies and projects emerge daily. These \ac{AI} technologies are essentially improved \ac{ML} models fine-tuned for specific tasks, learning rules and patterns from data to perform tasks like image classification. Large models like ChatGPT are often trained on internet data, submitted by humans, raising the question: if \ac{ML} models are trained with personal data, can they leak information about individuals? Researchers have shown that \ac{ML} models can be attacked. \Cref{fig:inverting_gradient_information_attack} shows one such attack, reconstructing training images from gradient information~\cite{Geiping_Bauermeister_Dröge_Moeller_2020}.
These attacks can affect privacy and lead to potentially harmful consequences, if an attacker obtains a victim’s sensitive personal information. Data breaches also generally violate GDPR~\cite{EuropeanParliament2016a} and may result in high fines, as seen with Cosmote Mobile Telecommunications, which paid €6,000,000 after a data breach~\cite{online-greece-Telecom}.

\ac{PPML} aims to prevent \ac{ML} models from leaking sensitive information by increasing the effective application of privacy guarantees~\cite{Xu_Baracaldo_Joshi_2021}. A common approach is introducing \ac{DP} to non-private \ac{ML} models, turning them into private ones~\cite{Al-Rubaie_Chang_2019}. The main drawback is the trade-off between model utility and privacy guarantees, requiring significant system tweaking to balance both interests.
Research has explored private learning approaches and \ac{PPML}, but less is known about the impact of dataset characteristics on \ac{ML} attacks or defenses~\cite{Xu_Baracaldo_Joshi_2021}. Understanding these effects can accelerate efforts by guiding system tuning~\cite{Truex_Liu_Gursoy_Yu_Wei_2021,Shokri_Stronati_Song_Shmatikov_2017,Carlini_Chien_Nasr_Song_Terzis_Tramer_2022,langePrivacyPracticePrivate2023}.

We investigate the influence of different dataset characteristics on the behavior of private and non-private image classification models.
We measure this by training models on various differing datasets and attacking them with a state-of-the-art \ac{MIA}, which infers whether a data sample was part of the model’s training set.
We focus on \ac{CNN}-based image classification models, a common \ac{MLaaS}, since previous \ac{PPML} research also underlines that image datasets and \ac{CNN}s are prone to attacks~\cite{Carlini_Chien_Nasr_Song_Terzis_Tramer_2022,Shokri_Stronati_Song_Shmatikov_2017,Truex_Liu_Gursoy_Yu_Wei_2021,Shamsabadi_Papernot_2022}.

We extend previous work that have looked at the relation of dataset characteristics and \ac{ML}s, by considering new metrics and private learning scenarios.
This can help assess whether applying \ac{DP} is worthwhile for a given dataset and what data considerations are needed for secure yet useful \ac{ML} applications.
We thereby take a more holistic view of the impact of dataset characteristics on both model utility and vulnerability to provide data-centric best practices for building private \ac{ML} models.
We discuss our final results and derived recommendations in \cref{sec:discussion}.
Our key findings resolve around an a priori analysis of vulnerability based on our dataset metrics.
We further find that our models effectively mitigate most of the \ac{MIA} threat with just a modest DP guarantee, achieving a more practical utility-privacy trade-off at low risk.

This work is structured continuing in \cref{sec:background}, which presents essential concepts. \Cref{sec:related} then reviews related work, while \cref{sec:exp} details our experiments, with \cref{sec:eval} discussing their results. \Cref{sec:conclusion} finally summarizes contributions and suggests future research directions.





\section{Background}
\label{sec:background}

This section provides an understanding of essential methods relevant to the experiments.

\subsection{Creating Private ML Models}
\label{sec:DP-SGD}

    \Citet{Abadi_Chu_Goodfellow_McMahan_Mironov_Talwar_Zhang_2016} introduced Differentially Private Stochastic Gradient Descent (DP-SGD), a modification of the SGD optimizer for \ac{DP}. DP-SGD limits privacy loss by clipping gradients and adding Gaussian noise. The key parameter for private training is the privacy budget ($\varepsilon$), which sets the wanted DP privacy level and allows a translation to the needed clipping and noise parameters.
    In their work~\citet{Ponomareva_Hazimeh_Kurakin_Xu_Denison_McMahan_Vassilvitskii_Chien_Thakurta_2023} propose a guide for creating differentially-private machine learning applications and give an evaluation of privacy budget needs.
    They constitute that an $\varepsilon \leq 1$ provides strong formal privacy guarantees, while more realistic privacy guarantees more likely use an $\varepsilon \leq 10$, which they believe still provides a reasonable utility-privacy trade-off.
    They further define $\varepsilon > 10$ as giving just weak to no formal privacy guarantees.
    The authors argue these applications may still be protected against attacks, but without a real formal guarantee.

\subsection{Likelihood Ratio Attack (LiRA)}
\label{sec:Likelihood Ratio Attack}
    
    We evaluate model vulnerability using the \ac{LiRA}~\cite{Carlini_Chien_Nasr_Song_Terzis_Tramer_2022}, a Membership Inference Attack (\ac{MIA})~\cite{Shokri_Stronati_Song_Shmatikov_2017}.
    \ac{MIA}s try to determine if a sample was part of the target model’s training set by exploiting its confidence scores.
    The \ac{LiRA} analyzes confidence distributions of shadow models, which involves training $N$ shadow models on random samples, half including the query sample $(x,y)$. These are IN and OUT models, respectively. Their confidence outputs are fitted to Gaussian distributions and the target model $f$ is then queried on $(x,y)$. A likelihood ratio test is performed comparing the target model’s output to the IN/OUT distributions, producing a \ac{LiRA} score indicating the membership probability. 
    
\section{Related Work}
\label{sec:related}
    Prior work has investigated the influence of dataset characteristics on the vulnerability of \ac{ML} models to \ac{MIA}s. \Citet{Shokri_Stronati_Song_Shmatikov_2017} found that models trained on datasets with more classes are more vulnerable, as the model must extract more discriminative features and thus retains more information about the training data. They also showed that classes with fewer samples are more susceptible to attack.
    Building on this, \citet{Truex_Liu_Gursoy_Yu_Wei_2021} demonstrated that models trained on datasets with larger class sizes are more vulnerable to \ac{MIA}s. They also found that datasets with higher in-class feature vector standard deviations lead to more accurate attacks, as outlier samples have a greater influence on model training. Additionally, they showed that creating minority classes artificially increases the vulnerability of those classes in a previous work~\cite{Truex_Liu_Gursoy_Wei_Yu_2019}.
    \citet{Tonni_Vatsalan_Farokhi_Kaafar_Lu_Tangari_2020} examined the effects of dataset size and class balance on MIA vulnerability in non-private settings. They found that smaller datasets are more susceptible to attack due to overfitting, and that minority classes are more vulnerable than majority classes. They also observed that higher data entropy decreases attack accuracy, as greater randomness makes it harder to infer membership information.
    \citet{dealcala2024comprehensive} recently looked into some other factors like batch size, dropout, and number of epochs.
    Overall, this prior work highlights the importance of dataset characteristics in determining the vulnerability of machine learning models to \ac{MIA}s. However, previous work in this area has mostly focused on the characteristics that increase non-private model vulnerability, but has not evaluated the effects in private learning scenarios.
    
    With their review, \citet{Ponomareva_Hazimeh_Kurakin_Xu_Denison_McMahan_Vassilvitskii_Chien_Thakurta_2023} offer guidance on successfully training a private ML model. However, they are only focused on tweaking the actual ML training process for better results and miss out on including the underlying training data to create a complete and comprehensive picture of an ML application.
    Thus, while related work looked into general training and some data-related parameters, there is no comprehensive evaluation of a wider range of dataset characteristics, including both dataset-level (e.g., class size, class count, imbalance) and data-level (e.g., information density, color, class separability) properties. We fill this gap by providing a comparative analysis of the effects of image dataset characteristics across different privacy budgets ($\varepsilon = \infty$, $\varepsilon = 30$, $\varepsilon = 1$), providing insights into how their influence on model behavior plays changing roles with stronger privacy guarantees.
 
\section{Experimental Setup}
\label{sec:exp}

\begin{figure}[t]
    \begin{center}
        \includegraphics[width=\textwidth]{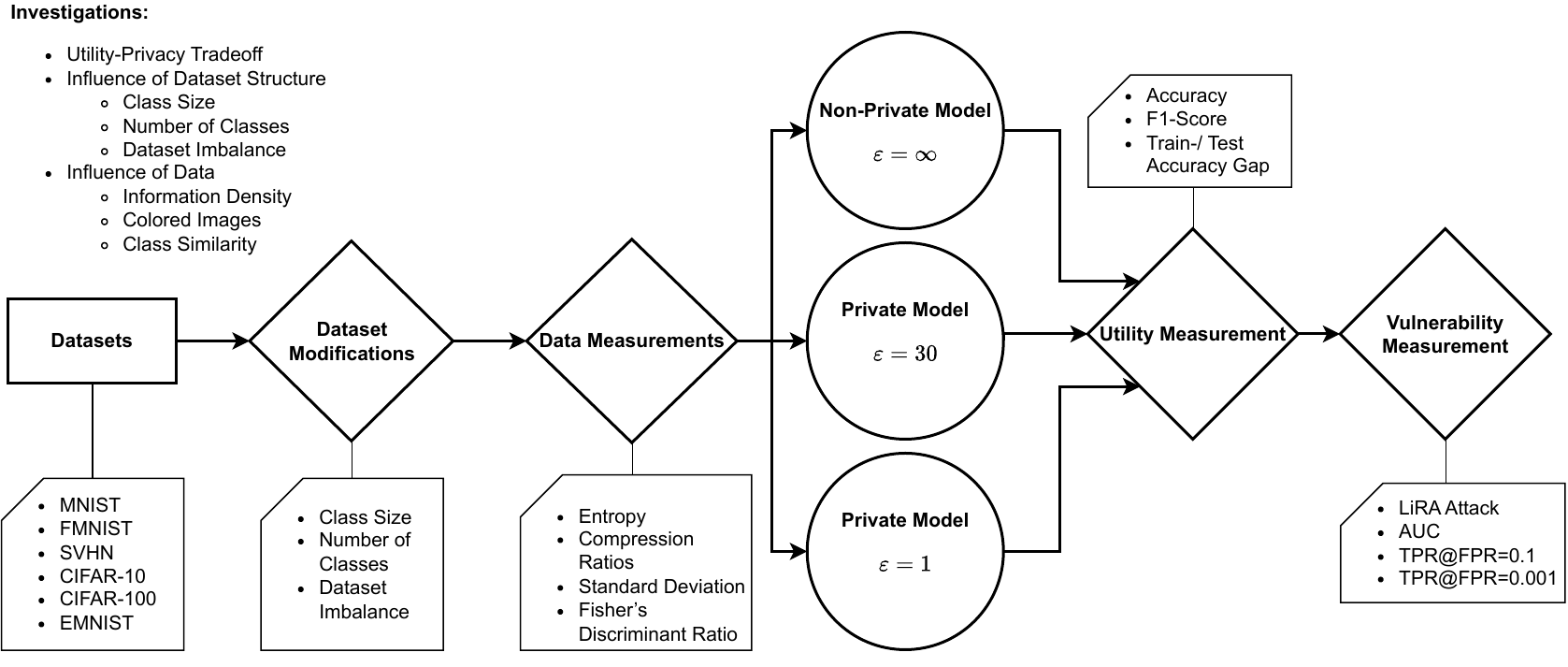}
    \end{center}
    \caption[Experiment Procedure Schematic]{Illustration of aspects and procedures in this work's experiments.}
    \label{fig:experiment_procedure}
\end{figure}

The experiments aim to determine how different dataset characteristics affect the utility and privacy of private \ac{ML} models. These characteristics include dataset size, number of classes, class balance, and properties like information density of images, color and grayscale influence, and class similarity. The results provide guidance for practitioners and researchers working with private \ac{ML} models.
A schematic overview of the experiments’ procedure is depicted in \cref{fig:experiment_procedure} and in this section, we give an overview on how we implemented this process. The experimental procedure has the following three steps, repeated with minor parameter changes to examine different aspects:
\begin{enumerate}
    \item \textbf{Non-Private Model Creation}: Train non-private models on different datasets for image classification. The models share the same architecture but are trained on datasets differing in one characteristic (e.g., class size). This allows for comparing utility and privacy based on that characteristic.

    \item \textbf{Private Model Training}: Convert non-private models to private models using a private learning algorithm to ensure \ac{DP}. Train models with high and low privacy budgets to assess utility loss as a trade-off for privacy. Compare prediction results to evaluate the dataset characteristics' influence on private model utility.

    \item \textbf{Model Attacks}: Attack the private and non-private models using \ac{MIA}s. Evaluate and compare the privacy of models with the same architecture but different training datasets. The attacks provide vulnerability metrics to understand how dataset characteristics affect private and non-private \ac{ML} models' proneness to attacks.
\end{enumerate}

\subsection{Environment}

    All experiments run on a computing cluster using an NVIDIA Tesla V100 GPU.
    The software stack utilizes Python 3.9 with Google's \ac{ML} library \textit{TensorFlow}~\cite{tensorflow2015-whitepaper} and the accompanying Keras framework for model training.
    In addition, the \textit{TensorFlow Privacy}\footnote{\url{https://github.com/tensorflow/privacy}} library is used to create private models and also offers capabilities for attacking models with \ac{MIA}s.
    To further emphasize the experiment's reproducibility, a fixed random seed value of 42 is used for all instances of random initialization, shuffling, sorting, or any other random methods executed on the datasets.
    Finally, reference code for all experiments is available from our repo at \url{https://github.com/luckyos-code/dataset-analysis-ppml}.

\subsection{Datasets}
\label{sec:Datasets}

\begin{figure}[t]
        \centering
        \begin{subfigure}[t]{0.32\textwidth}
            \includegraphics[trim= 0 15cm 0 0, clip, width=\textwidth]{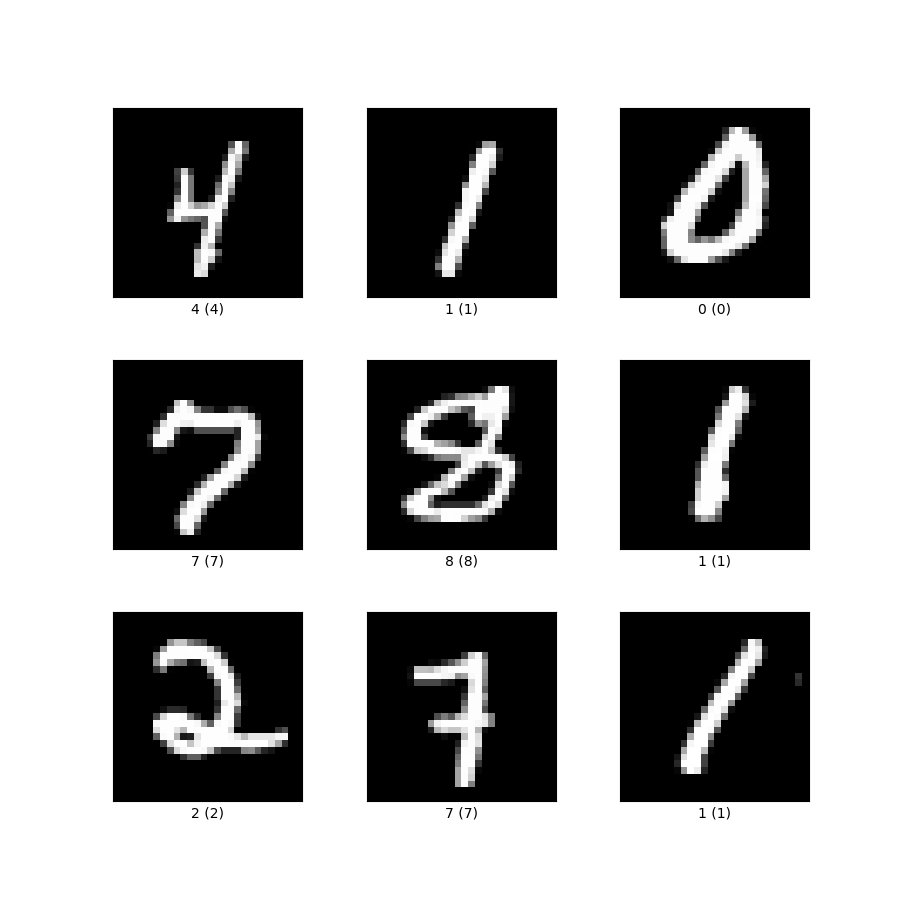}
            \caption{
               MNIST dataset.}
            \label{fig:mnist_example}
        \end{subfigure}
        \begin{subfigure}[t]{0.32\textwidth}
            \includegraphics[trim= 0 15cm 0 0, clip, width=\textwidth]{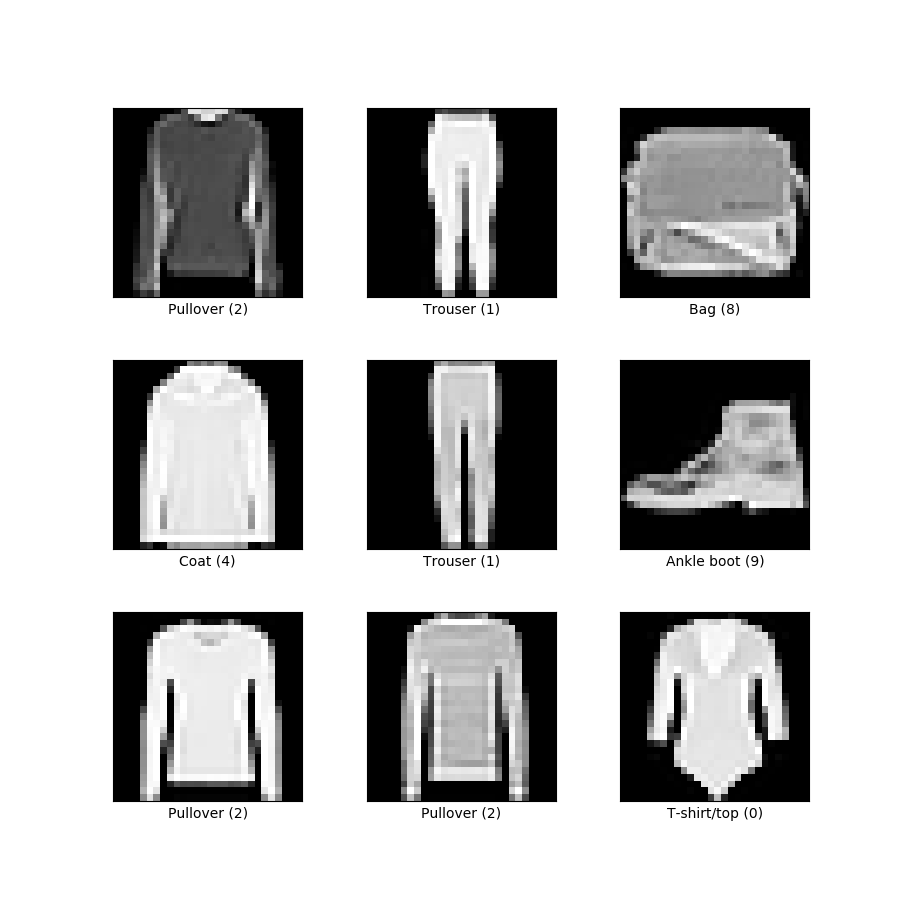}
            \caption{FMNIST dataset.}
            \label{fig:fashion_mnist_example}
        \end{subfigure}
        \begin{subfigure}[t]{0.32\textwidth}
            \includegraphics[trim= 0 15cm 0 0, clip, width=\textwidth]{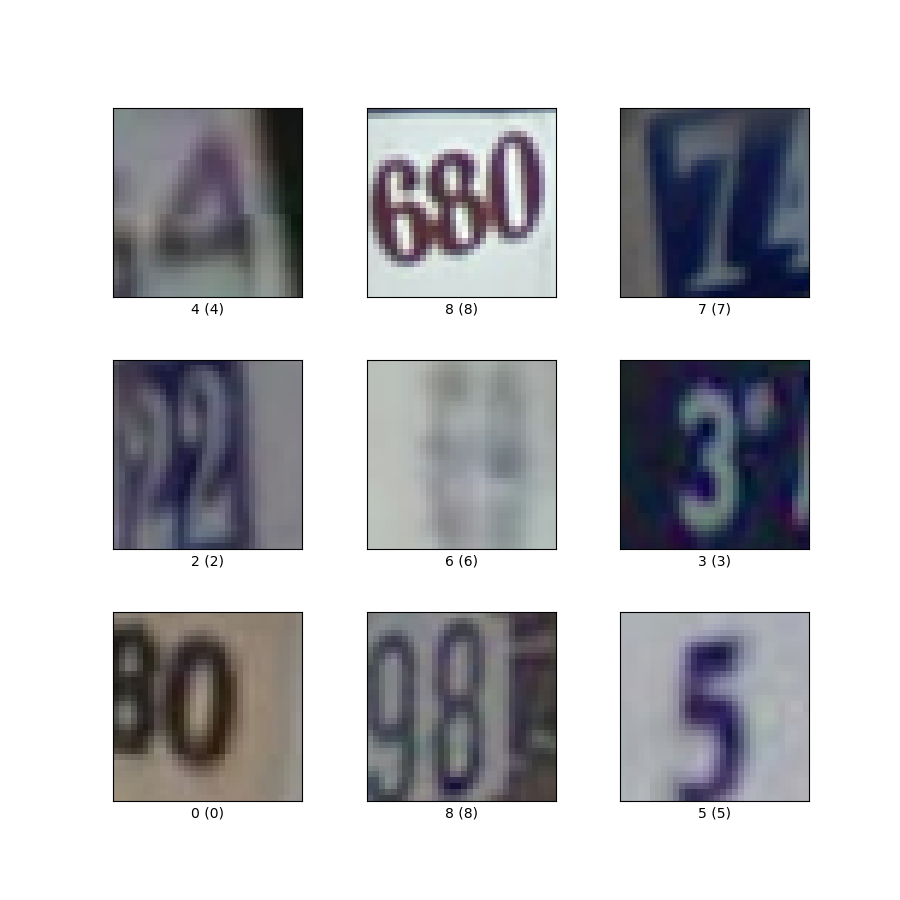}
            \caption{SVHN dataset.}
            \label{fig:svhn_example}
        \end{subfigure}
        \begin{subfigure}[t]{0.32\textwidth}
            \includegraphics[trim= 0 15cm 0 0, clip, width=\textwidth]{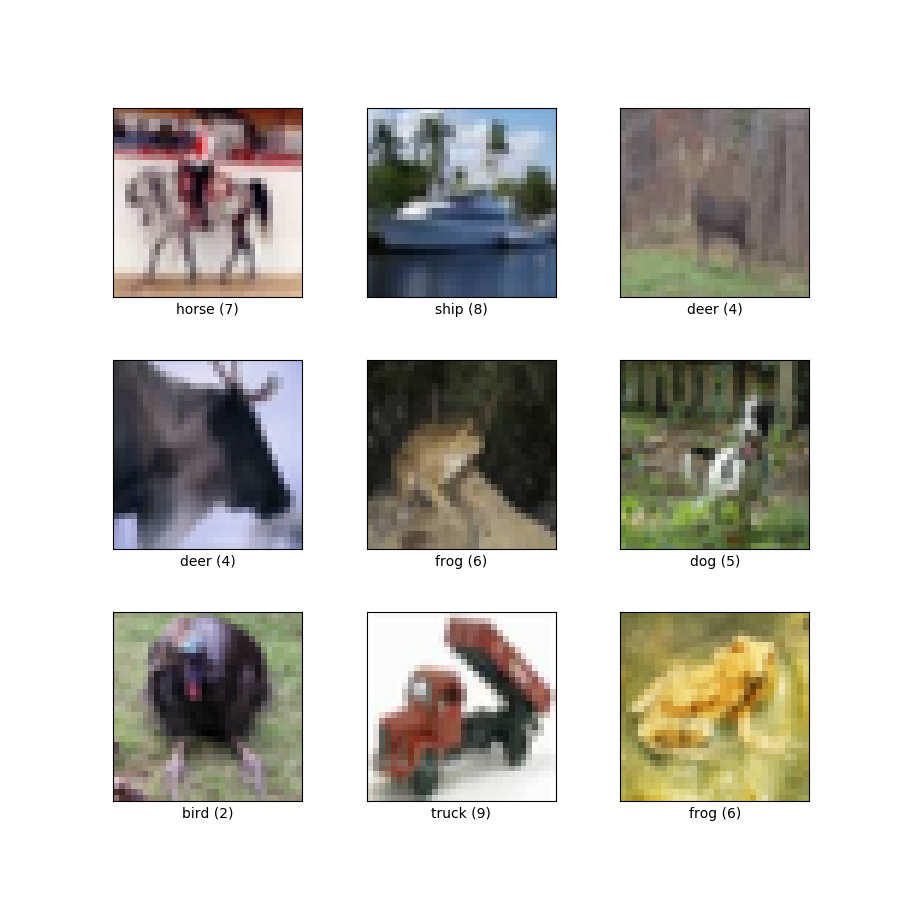}
            \caption{CIFAR-10 dataset.}
            \label{fig:cifar10_example}
        \end{subfigure}
        \begin{subfigure}[t]{0.32\textwidth}
            \includegraphics[trim= 0 14.8cm 0 0, clip, width=\textwidth]{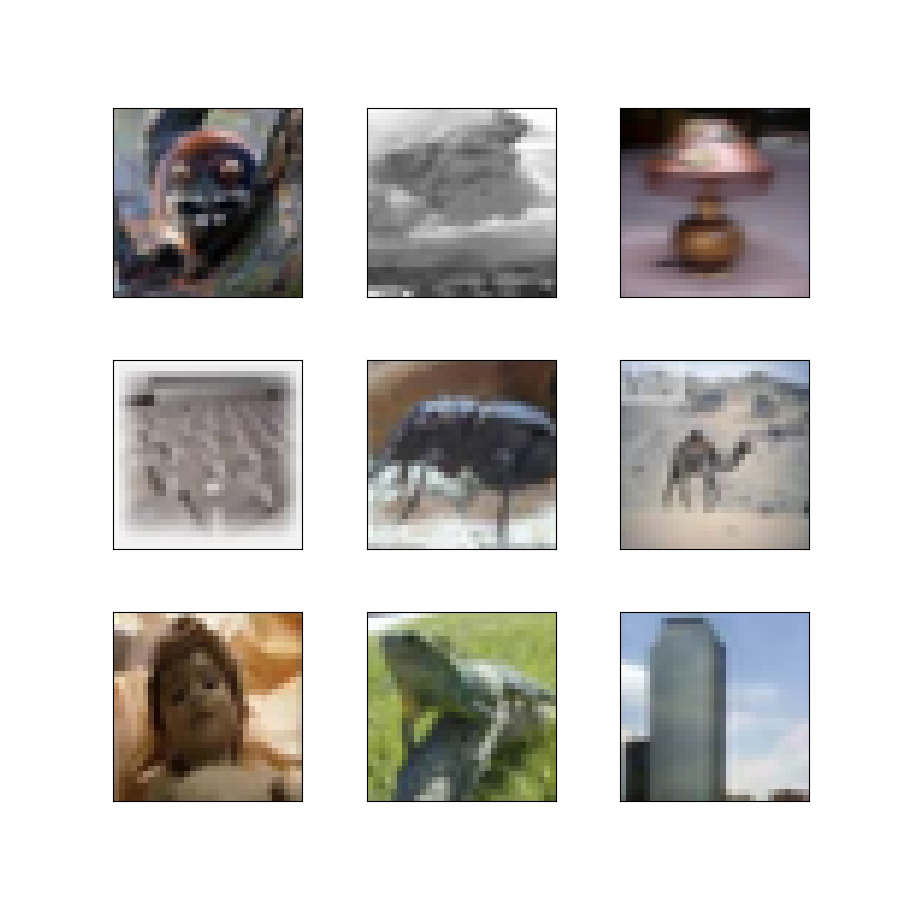}
            \caption{CIFAR-100 dataset.}
            \label{fig:cifar100_example}
        \end{subfigure}
        \begin{subfigure}[t]{0.32\textwidth}
            \includegraphics[trim= 0 15cm 0 0, clip, width=\textwidth]{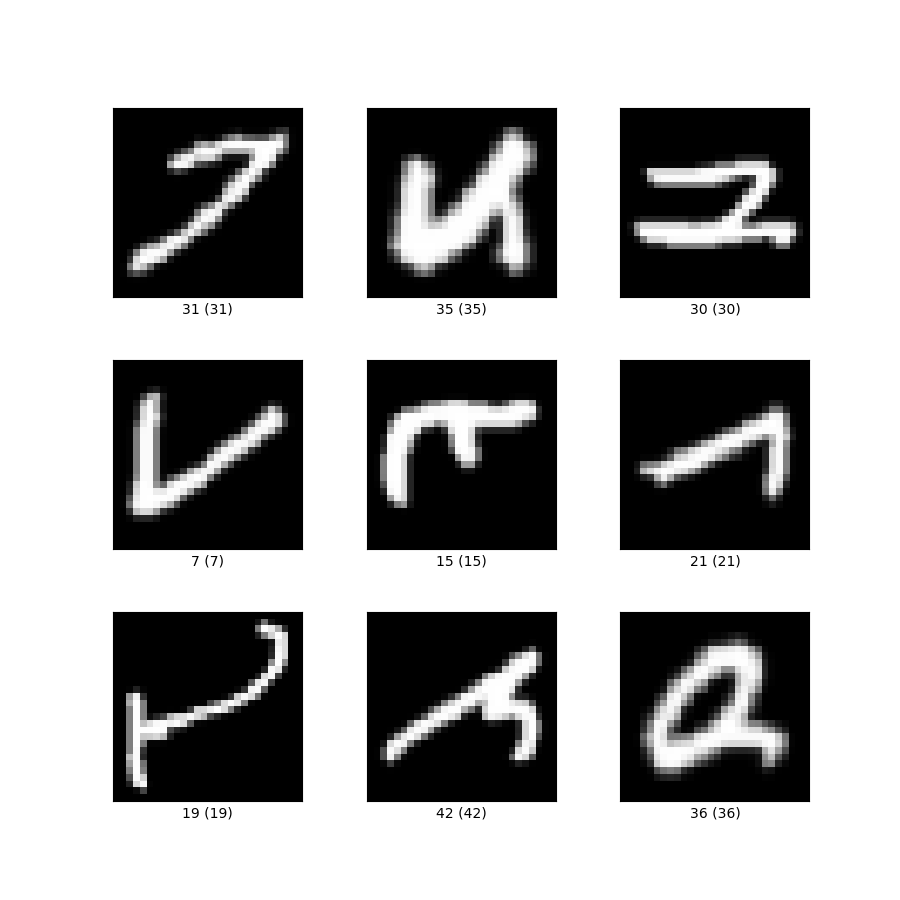}
            \caption{EMNIST dataset.}
            \label{fig:emnist_example}
        \end{subfigure}
        \caption{Visual representation of random samples from the studied image datasets.}
        \label{fig:dataset-visual}
    \end{figure}

    This section introduces the datasets used in our experiments and analyses, highlighting their key characteristics.
    All the datasets are image datasets commonly used for multi-class image classification tasks.
    While they do not directly contain sensitive information, they are among the most important benchmarking datasets in privacy-preserving machine learning (PPML) research, offering a diverse range of characteristics for our study~\cite{Abadi_Chu_Goodfellow_McMahan_Mironov_Talwar_Zhang_2016, DBLP:conf/iclr/PapernotSMRTE18, boenisch2024have, konevcny2016federated}.
    In \Cref{fig:dataset-visual} we provide sample images from each dataset.

        \textbf{MNIST.} The \ac{MNIST} dataset~\cite{Lecun_Bottou_Bengio_Haffner_1998} is known for its simple structure and small image size and was originally created for a document recognition task.
        It contains 10 classes representing handwritten digits with labels from 0 to 9.
        The images as shown in \cref{fig:mnist_example}, are in grayscale and have a size of 28×28px.
        In total there are 70,000 images in the dataset, split up in 60,000 images for training and 10,000 images for testing.
        The \ac{MNIST} dataset is often used as a benchmark for (private) image classification tasks~\cite{Abadi_Chu_Goodfellow_McMahan_Mironov_Talwar_Zhang_2016, Uniyal_Naidu_Kotti_Singh_Kenfack_Mireshghallah_Trask_2021, langePrivacyPracticePrivate2023} or to analyze the different learning approaches~\cite{Perez_Wang_2017, Kingma_Ba_2014, Srivastava_Hinton_Krizhevsky_Sutskever_Salakhutdinov_2014}---making it a perfect basic candidate for our experiments.
    
    
        \textbf{FMNIST.} The \ac{MNIST} dataset is regularly criticized for its simplicity and age, since \ac{ML} developed much since its initial release.
        Therefore, a more complex benchmarking dataset was needed, resulting in the \ac{FMNIST} dataset~\cite{DBLP:journals/corr/abs-1708-07747}.
        The dataset shares the same features as the \ac{MNIST} dataset in all but the more complex images depicting products from an online fashion store and have a much wider variety of shapes and textures than the \ac{MNIST} dataset, as is shown in \cref{fig:fashion_mnist_example}.
        Thanks to its matching structure, \ac{FMNIST} is a suitable choice for a direct comparsion to the \ac{MNIST} dataset.
        Since it contains a wider variety in patterns and textures it can be considered as a more complex task.

    
        \textbf{SVHN.} The \ac{SVHN} dataset~\cite{netzer2011reading} contains cropped RGB images of house numbers from real Google Street View\footnote{\url{https://www.google.com/streetview/}} images, with examples given in \cref{fig:svhn_example}.
        It has 10 classes, each representing a digit, and features about 73,000 training and 26,000 test images with 32×32px in size.
        The task is comparable to \ac{MNIST}, but in contrast to the \ac{MNIST} or \ac{FMNIST} datasets, this dataset now contains RGB color images and the recognition task is again harder than the \ac{MNIST} task~\cite{netzer2011reading}.

    
        \textbf{CIFAR-10.} With the \ac{CIFAR}-10~\cite{Krizhevsky09learningmultiple}, we introduce another well-known RGB image classification dataset.
        It again consists of 10 classes and only a small image size of 32x32px, however, the depicted objects in these classes are real-life objects like horses, ships, deer, trucks, etc., shown in colorful images.
        Each class contains exactly 6,000 images and the resulting balanced dataset provides 50,000 samples for training and 10,000 for testing.
        Due to its color images, complex patterns, and non-uniform backgrounds, the multi-class image classification on \ac{CIFAR}-10 is considered as an even more difficult task than with the \ac{FMNIST} dataset~\cite{Kowsari_Heidarysafa_Brown_Meimandi_Barnes_2018}.
        An excerpt of a few images from the dataset are shown in \cref{fig:cifar10_example}.
        While featuring an increasingly complex task, the dataset again keeps a similar image size and class count as the \ac{MNIST} and \ac{FMNIST} datasets, which is an important aspect for explainable results.
    
    
        \textbf{CIFAR-100.} The \ac{CIFAR}-100 dataset~\cite{Krizhevsky09learningmultiple} is an extension to the \ac{CIFAR}-10 dataset and while they both contain RGB color images with a size of 32×32px, as seen in \cref{fig:cifar100_example}, the \ac{CIFAR}-100 instead now consists of 100 classes with 600 images per class.
        This leads to an $10\times$ increase in classes and an according decrease in samples per class.
        Keeping all factors similar to the \ac{CIFAR}-10 and just drastically changing the amount of classes allows studying the influence of the number of classes on the model's performance and vulnerability.
    
    
        \textbf{EMNIST.} The \ac{EMNIST} dataset~\cite{Cohen_Afshar_Tapson_van} is an extension to the \ac{MNIST} dataset that contains the complete NIST Special database 19~\cite{grother1995nist}.
        This is the database \ac{MNIST} was derived from and means that the \ac{EMNIST} contains exactly the same type of images.
        However, instead of only using handwritten digits, \ac{EMNIST} additionally features uppercase and lowercase handwritten letters, with examples given in \cref{fig:emnist_example}.
        The dataset comes in different variations and we decided on the balanced version of the By\_Merge variation~\cite{Cohen_Afshar_Tapson_van}, which results in 47 classes for 131,600 images.
        With similar image complexity as the \ac{MNIST} dataset but providing more classes, \ac{EMNIST} fits the same spot as the \ac{CIFAR}-100 dataset enabling an isolated study regarding the class count.

\subsection{Measuring Model Utility and Vulnerability}
\label{sec:metrics}
    
    Measuring model utility helps assess the impact of dataset properties on both non-private and private learning. It is essential for balancing model privacy and utility. We use accuracy and F1-score to evaluate our multi-class classification tasks, with F1-score being preferred for evaluating unbalanced datasets~\cite{Vakili_Ghamsari_Rezaei_2020, Sokolova_Japkowicz_Szpakowicz_2006,Hripcsak_2005}.
    Model accuracy is also used to calculate the train-test accuracy gap, indicating overfitting~\cite{Hardt_Recht_Singer_2015}.
    
    Practically measuring model privacy is crucial, as our theoretical \ac{DP}-guarantees just provide an upper bound on possible information leakage and should thus not be taken as the single true indicator of privacy risk~\cite{Rahman2018MembershipIA}. \ac{MIA}s can help evaluate this risk by determining a practical risk level through actual attacks, which can then be connected to our different privacy levels and datasets~\cite{malek2021antipodes}. We use the proposed offline version of the white-box \ac{LiRA} \ac{MIA}~\cite{Carlini_Chien_Nasr_Song_Terzis_Tramer_2022} with 32 shadow models. The offline attack eliminates the need for the IN shadow models described in \cref{sec:Likelihood Ratio Attack}.
    
    Three metrics measure the attack's effectiveness: \ac{ROC}-\ac{AUC}, \ac{TPR} @ 0.1 \ac{FPR}, and \ac{TPR}@0.001 \ac{FPR}, which were suggested by \citet{Carlini_Chien_Nasr_Song_Terzis_Tramer_2022}.
    They promote measuring the \ac{TPR} at a low fixed \ac{FPR}, since a low \ac{FPR} rate is most important for an attacker.
    In general, a \ac{ROC} curve visualizes the relationship between an attack's \ac{TPR} and \ac{FPR}, which is then captured across all FPR values using the \ac{AUC}.
    All three metrics have baseline values constituting a randomly guessing attacker, which would translate to perfect privacy.
    The baseline for the \ac{AUC} metric is $0.5$, while the baselines for TPR are $0.1$ and $0.001$, respectively.
    Finally, each shadow model gives a single attack result for its part of the dataset and we therefore give average results over all shadow models.

\subsection{Experiments}
    In this part, we detail our evaluation strategy for relevant dataset characteristics.
    A general overview of the experiments is given in \cref{fig:experiment_procedure}.
    The experiment settings are intended to provide information to help researchers and practitioners working with \ac{PPML}.
    The results should help them in designing a secure private \ac{ML} model faster by considering the influence of different aspects of a dataset in the process.
    The experiments explore which dataset characteristics increase or decrease the vulnerability and utility of private and non-private \ac{ML} models, which is evaluated by the metrics described in \cref{sec:metrics}.
    For the private model experiments, we further distinguish between two privacy scenarios using different privacy budgets ($\varepsilon=\{1,30\}$).
    
    The dataset characteristics are divided into two different levels depending on what kind of modifcations are applied.
    The first set of characteristics considers the \textit{dataset-level}, which focuses on the overall dataset structure with statistics like number of samples, class count and class imbalance.
    The second level consists of \textit{data-level} modifications, which are characteristics that e.g., measure data complexity, the influence of color information, and class separability.
    While we can generally change dataset-level factors by using fitting slices of the dataset, modifying most underlying data-level characteristics like data complexity is much more difficult.
    To avoid this problem, we instead use a selection of differing datasets as presented in \cref{sec:Datasets} to better evaluate these data-level characteristics.

    For each setting, we define a set of modifications with changing parameters that we apply only to a dataset's training split, leaving the test split untouched for our evaluating metrics.
    For each modified dataset version, we train a non-private and two private models with our different privacy budgets $\varepsilon=1$ and $\varepsilon=30$.
    On these models we then measure the impact of the varying dataset setups for each modification and privacy level, by first calculating their utility scores and then by attacking them with the \ac{LiRA} attack.
    In the following, we provide information on our used modifications.
    
    \subsubsection{Dataset-Level Investigations}
    \label{subsec:Dataset Level Investigations}
    
        The dataset-level investigations are a series of experiments to assess how overall dataset statistics influence the private learning of \ac{ML} models.
        
        \textbf{Class size.}
            The first dataset modification introduces a reduction and normalization of the class size, i.e., the number of samples per class.
            It reduces the number of samples in each class to a fixed value, which in our experiments is called $c$.
            This reduction is done by randomly removing samples per class until the specified number of samples per class is reached.
            This also means, that we achieve a perfect class balance, since all classes are equal in their sample counts.
            The datasets used in this experiment are \ac{MNIST}, \ac{FMNIST}, \ac{SVHN} and \ac{CIFAR}-10, for which we run several experiments with different class sizes.
            In total, there are eight class sizes, which successively reduce the size to 5000, 4000, 3000, 2000, 1000, 500, 100 and 50 samples per class.
            The biggest size of $c=5000$ is the largest common class size across all datasets, acting as the baseline experiment.
            This amount is enough to see if bigger classes reduce the privacy risk due to more available data.
        
        \textbf{Class count.} 
            The class count modification changes the number of classes in a dataset by purposefully reducing them, which is done by deleting all samples of a set number of existing classes.
            The resulting class count is denoted in the parameter $n$ and to make the deletion process deterministic for all datasets, first, all available class labels are retrieved and placed in an alphanumerically ordered list.
            From this ordered list, only the first $n$ labels are kept and the rest are discarded together with their respective samples in the dataset.
            
            This experiment is divided into two sub-parts depending on the original dataset classes available.
            The first part, we use the datasets with fewer classes, namely the \ac{MNIST}, \ac{FMNIST}, \ac{SVHN} and \ac{CIFAR}-10.
            Their class counts are each reduced from 10 down to 3, which is the border for still keeping a multi-class classification task.
            The second part aims at the \ac{CIFAR}-100 and \ac{EMNIST}, which consist of many more classes.
            The baseline is based on \ac{EMNIST}, which contains 47 classes, while CIFAR-100 has even more at 100 classes.
            With this, we reduce the classes over a larger range than before to generate further insights.
            We start at 47 classes, which is the baseline as it is the maximum for \ac{EMNIST}, and reduce the class size in steps of five until it reaches three classes.
        
        \textbf{Class imbalance.}
            Class imbalance describes the issue of a skewed data distribution, meaning that some classes contain more samples than others, which results in minority and majority classes.
            While minority classes are classes that are underrepresented, by having much lesser samples, majority classes are overrepresented and have much more samples than the minority classes.
            In this operation, we remove data samples to create an artificial class imbalance.
            For this we introduce the imbalance factor $i$, which we keep in a $[0,1]$ range.
            Here, $i = 0$ means that there is no class imbalance, and $i = 1$ translates to the highest possible class imbalance.
            
            Furthermore, this modification works in two modes.
            Both modes use the same datasets, which are \ac{MNIST}, \ac{FMNIST}, \ac{SVHN} and \ac{CIFAR}-10, with five values for $i$ at 0.0 (baseline), 0.1, 0.3, 0.6, and 0.9.
            Before applying the imbalance modification, we reduce all the class sizes to 5000 to start with perfectly balanced and equal-sized datasets.

\begin{figure}[t]
                \centering
                \begin{subfigure}[c][][c]{0.245\textwidth}
                    \includegraphics[width=\textwidth]{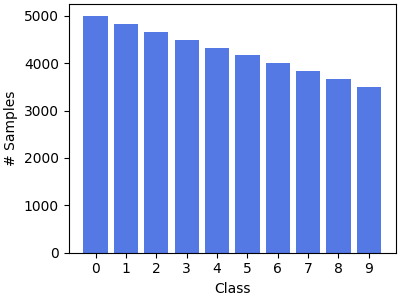}
                    \caption{$i=0.3$, linear.}
                    \label{fig:imbalance_experiment_class_distribution_a}
                \end{subfigure}
                \begin{subfigure}[c][][c]{0.245\textwidth}
                    \includegraphics[width=\textwidth]{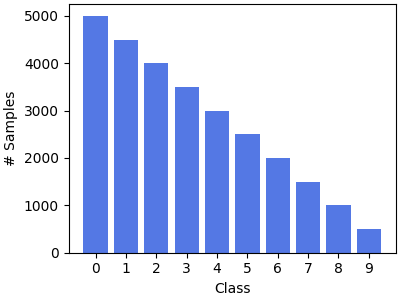}
                    \caption{$i=0.9$, linear.}
                    \label{fig:imbalance_experiment_class_distribution_b}
                \end{subfigure}
                \begin{subfigure}[c][][c]{0.245\textwidth}
                    \includegraphics[width=\textwidth]{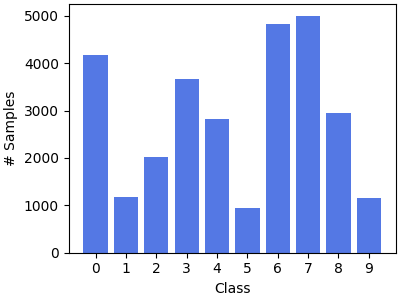}
                    \caption{$i=0.3$, normal.}
                    \label{fig:imbalance_experiment_class_distribution_c}
                \end{subfigure}
                \begin{subfigure}[c][][c]{0.245\textwidth}
                    \includegraphics[width=\textwidth]{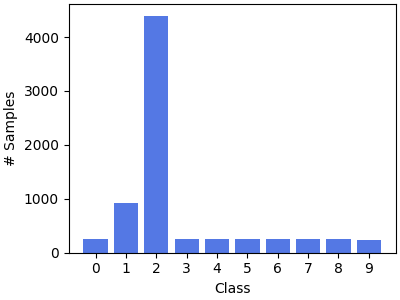}
                    \caption{$i=0.9$, normal.}
                    \label{fig:imbalance_experiment_class_distribution_d}
                \end{subfigure}
                \caption[Dataset Class Distribution for Dataset Imbalance Modification in Linear Mode]{
                    Visualization of the dataset class distribution after applying the dataset imbalance modification in linear and normal mode with varying imbalance factors $i=0.3$ and $i=0.9$.}
                \label{fig:imbalance_experiment_class_distribution}
            \end{figure}

            The first mode applies imbalance in \textit{linear} mode, which means that the classes of the dataset decrease linearly in size, and no two classes have the same size.
            This is done by first sorting all classes by their current class size.
            Then the smallest class size is multiplied by $1-i$, which is the final new class size for the smallest class in the dataset.
            Now the class size values between the first and largest class, and the last and smallest class are set linearly ascending with equal intervals.
            The resulting distribution of samples in linear mode for $i=\{0.3,0.9\}$ is visualized in \cref{fig:imbalance_experiment_class_distribution_a,fig:imbalance_experiment_class_distribution_b}.
            
            The second is the \textit{normal} mode, which creates imbalance by randomly removing samples based on a normal distribution, with its mean as $\text{mean}=1-i$ and the standard deviation as $\text{std}=i$.
            We further clip the drawn values at $1.0$ and $0.05$.
            The sampled values are multiplied by the current class size, resulting in a new and possibly reduced size.
            The modified distribution in normal mode is visualized in \cref{fig:imbalance_experiment_class_distribution_c,fig:imbalance_experiment_class_distribution_d}.
            The visualization shows the randomness of normal mode imbalance compared to linear mode.
            
    \subsubsection{Data-Level Investigations}
    \label{subsec:Data Level Experiments}
    
        While the dataset-level experiments (see \cref{subsec:Dataset Level Investigations}) investigate the model behavior in relation to the structural properties of the dataset, the data-level experiments observe how the data itself influences the private and non-private \ac{ML} models.
        The focus of the data-level investigations is to find and compute quantifiable metrics that describe the image data itself.
        
        \textbf{Information density.}
            One way to describe the complexity of datasets is to measure the density of information in the data.
            Comparing the \ac{MNIST} and \ac{CIFAR}-10 images gives a general sense of what is meant by this characteristic.
            The \ac{MNIST} dataset consists of grayscale pixels forming simple structures, whereas the \ac{CIFAR}-10 dataset has a much wider variety of pixel brightness, complex textures, and patterns.
            In short, the \ac{CIFAR}-10 images appear more complex than the \ac{MNIST} images.
            We quantify image complexity by using two methods.
            The first method is to calculate the Shannon entropy~\cite{Shannon_1948}, which defines as a measure of randomness inside the data.
            A high entropy value indicates that there is more disorder or randomness in the data, while a lower entropy value means that the data is more ordered and predictable.
            The idea is that more information in a dataset translates to less uncertainty.
            We calculate the entropy $H$ image-wise as follows: $H(X) = - \sum_{x \in X} p(x) \log p(x)$.
            Where $X$ is the set of all pixel values in an image and $p(x)$ describes the probability that a pixel with value $x$ occurs.
            This equation works very well for grayscale images, however, calculating the entropy ($H$) of color images requires an extra step, in which we calculate an averaged entropy over the three color channels.

            The second measure of information density is the compression ratio.
            In~\cite{Yu_Winkler_2013}, the authors argue that Shannon's entropy is not a good measure of image complexity because it does not account for spatial structure.
            They therefore define a compression ratio measure, which is defined as: $CR = \frac{s(I)}{s(C(I))}$.
            Where $s(I)$ is the uncompressed image size in bytes and $s(C(I))$ is the compressed image size in bytes.
            This ratio is an indicator of how much the image $I$ can be compressed.
            The idea behind this measure is that noise patterns in images that do not contain much information, but consist of random patterns, cannot be compressed as efficiently as patterns that contain actual visual information.
            We use two different compression methods, a lossy and a lossless compression.
            A lossy compression method loses some image information during image compression, while a lossless compression method can compress images without losing any information.
            The lossy method used is the \ac{JPEG}~\cite{O_Brien_2005}, which utilizes visual features by extracting frequency information from the image.
            The second approach uses the lossless \ac{PNG} image compression~\cite{Boutell_1997}, which compared to the JPEG, works more on the pixel level of the image rather than the visual characteristics.
        
        \textbf{Color.}
            For experimenting with color, we take \ac{SVHN} and \ac{CIFAR}-10, once with their original color and once in a modified grayscale version.
            Three information channel (RGB) should provide more information than one channel of information (grayscale).
            To transform color images to grayscale, we multiply each RGB color channel (red, green, blue) by a specific weight representing the wavelength of the color.
            These values are then summed up to represent the grayscale value for that pixel.

        \textbf{Class similarity.}
            Class similarity describes the similarity of data samples within a class or between classes.
            The intuition behind analyzing class similarity is to roughly estimate task difficulty of multi-class classification problems.
            If the samples between classes are similar to each other, the classification task may become harder, since the rules needed to classify the data samples may be harder to learn for an \ac{ML} model.
            On the other hand, if many samples within a class are very similar, they are easier to classify as belonging to that class because they share more common attributes.
            Class similarity is analyzed using two different methods.
            The first method follows the work of~\citet{Truex_Liu_Gursoy_Yu_Wei_2021} and calculates the in-class feature vector \ac{STD} of the dataset.
            The motivation behind using the \ac{STD} is to calculate how much the samples within a group differ from each other on average, which is an indicator of how similar the data is.
            The second method of class similarity calculates the \ac{FDR}.
            This ratio measures the degree to which the classes of a dataset are separable.
            The \ac{FDR} is known from its use in Fisher's Linear Discriminant Analysis (LDA) as a measure to be maximized by the LDA procedure~\cite{Fisher_1936}.
            Basically, the \ac{FDR} is the ratio of the between-class variance to the within-class variance~\cite{Li_Wang_2014}.
            The motivation for analyzing class separability is that classes that are more similar to each other are harder to separate.
            Thus, a lower \ac{FDR} value for a dataset could indicate that this dataset is harder to classify than one with a higher \ac{FDR} value.

\subsection{Pre-processing}
\label{sec:Processing}

    Regarding pre-processing, we detail the needed modifications and augmentations applied to the training data to fit the model's input shape and improve utility. First, all datasets are split into train and test sets, using their provided standard splits to ensure representativeness and comparability with other work.
    With such differing inputs from our multiple datasets, we have to consolidate them into some common input space.
    Thus, all images are scaled before training to match the required 32×32 input shape for our model described in \cref{sec:model}. Datasets with smaller dimensions, such as MNIST (28×28), are upscaled accordingly. Next, 8-bit color image values are normalized to the [0, 1] range to improve convergence, particularly with \ac{ReLU} activation functions.
    For grayscale datasets like MNIST or Fashion-MNIST, which have one channel, two additional identical channels are added to create a 32×32×3 input shape to match the model's requirements, which has to classify images from all datasets.
    Data augmentation is a common tool for enhancing a model's generalization ability by combating overfitting. We apply horizontal flipping based on \citet{Carlini_Chien_Nasr_Song_Terzis_Tramer_2022}, where we randomly flip 50\% of images on the vertical axis.

\subsection{Model Architecture and Training}
\label{sec:model}

    Our chosen architecture is a \ac{CNN} similar to LeNet-5~\cite{Lecun_Bottou_Bengio_Haffner_1998} that we modified to better suit our tasks. 
    The model starts with a \textit{RandomFlip} layer, which randomly applies horizontal flips on the input to ensure the data augmentation from \cref{sec:Processing}. The rest of the model comprises three convolutional, one hidden dense, and one output dense layer, which are each connected by group normalization and max pooling layers.
    We decided on a basic but powerful architecture to allow maximum generalizability to \ac{CNN}s.
    Choosing differing or too complex models can distort the findings due to these newly introduced factors.
    %
    Following the guidelines from \citet{Ponomareva_Hazimeh_Kurakin_Xu_Denison_McMahan_Vassilvitskii_Chien_Thakurta_2023}, we performed a hyperparameter search. This results in using the Adam optimizer with a learning rate of $\alpha = 0.005$ and batch size of 256 over 30 epochs. For private training, we shift to DP-Adam, a differentially-private alternative, with microbatches of 256 and a clipping norm of $1.0$. 
    The noise parameters are dynamically determined according to our privacy budgets $\varepsilon=1$ and $\varepsilon=30$, since they also depend on the training data. For example, with 60,000 MNIST samples, batch size 256, and 30 epochs, the noise multipliers are $\sigma=0.431$ for $\varepsilon=30$ and $\sigma=1.626$ for $\varepsilon=1$.

\section{Evaluation}
\label{sec:eval}

This section presents the experimental results from \cref{sec:exp}, focusing on the dataset characteristics that influence the behavior of the \ac{ML} model, assessed through utility (accuracy, F1-score) and vulnerability (\ac{LiRA} \ac{MIA}).
We examine models at three different privacy levels: non-private models at $\varepsilon=\infty$, private models with $\varepsilon=30$, and $\varepsilon=1$.
We use multiple datasets, varying in the difficulty of their multi-class image classification tasks but having similar structures (see \cref{sec:Datasets}).

\subsection{Dataset-Level Results}
\label{subsec:Dataset Level Results}

This part investigates how dataset characteristics, such as class size, class count, and dataset imbalance, affect model utility and vulnerability.
    
    \textbf{Class size.}
        \begin{figure}[t]
            \centering
            \begin{subfigure}[c][][c]{0.328\textwidth}
                \includegraphics[width=\textwidth]{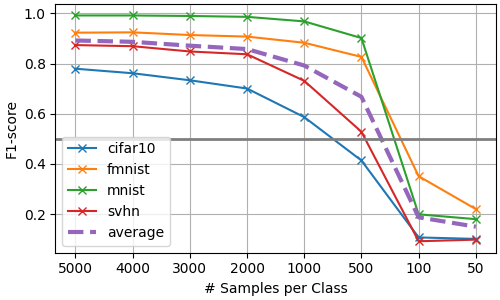}
                \caption{Non-private ($\varepsilon=\infty$) model.}
            \end{subfigure}
            \hfill
            \begin{subfigure}[c][][c]{0.328\textwidth}
                \includegraphics[width=\textwidth]{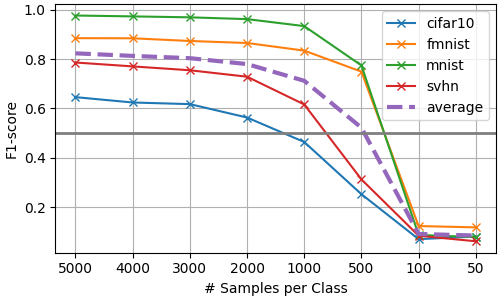}
                \caption{Private ($\varepsilon=30$) model.}
            \end{subfigure}
            \hfill
            \begin{subfigure}[c][][c]{0.328\textwidth}
                \includegraphics[width=\textwidth]{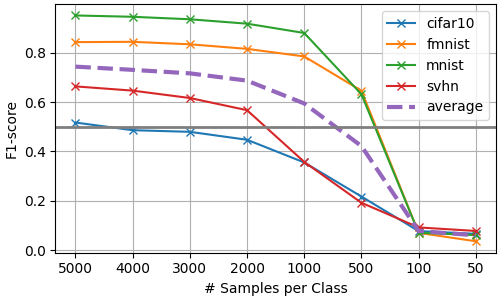}
                \caption{Private ($\varepsilon=1$) model.}
            \end{subfigure}
            \caption{
                F1-scores for non-private and private models on different datasets with modified class sizes.}
            \label{fig:class_size_f1_results}
        \end{figure}
        %
        %
        \begin{figure}[t]
            \centering
            \begin{subfigure}[c][][c]{0.3\textwidth}
                \includegraphics[width=\textwidth]{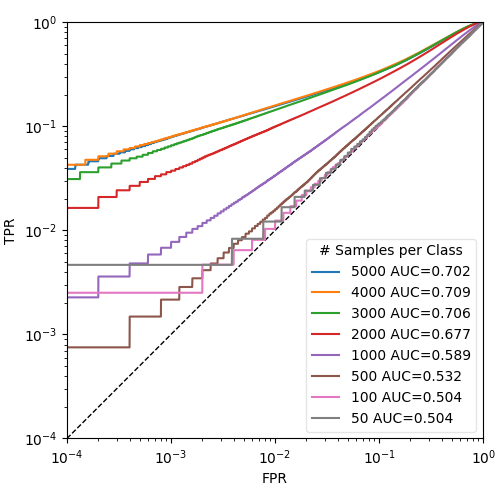}
                \caption{Non-private ($\varepsilon=\infty$) model.}
            \end{subfigure}
            \hfill
            \begin{subfigure}[c][][c]{0.3\textwidth}
                \includegraphics[width=\textwidth]{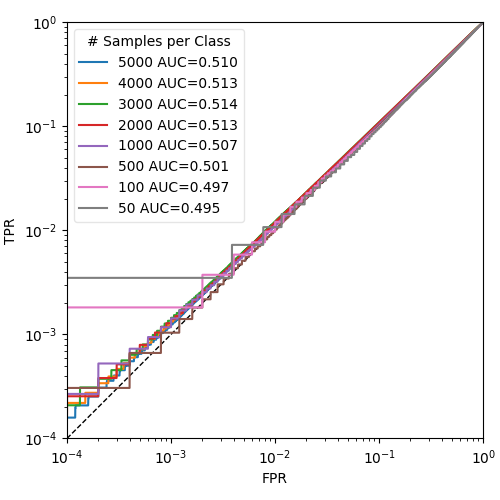}
                \caption{Private ($\varepsilon=30$) model.}
            \end{subfigure}
            \hfill
            \begin{subfigure}[c][][c]{0.3\textwidth}
                \includegraphics[width=\textwidth]{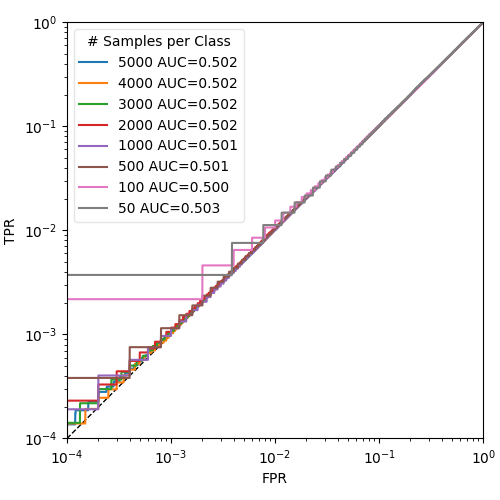}
                \caption{Private ($\varepsilon=1$) model.}
            \end{subfigure}
            \caption{
                Attack ROC curves for averaged non-private and private models with modified class sizes.}
            \label{fig:class_size_attack_roc_results}
        \end{figure}
        For evaluating the utility over our proposed class size reductions, we use \cref{fig:class_size_f1_results}, which shows the achieved F1-scores for each class size and privacy budget regarding different datasets.
        We generally observe that model utility decreases with fewer samples per class.
        For non-private models, the average F1-score dropped from $0.89$ (5000 samples/class) to $0.19$ (100 samples/class) and we find similar behavior for the private models.
        However, they additionally battle with reduced overall utility due to their utility-privacy trade-off.
        
        Our vulnerability test in \cref{fig:class_size_attack_roc_results} presents attack susceptibility across datasets using ROC curves.
        At 5000 samples per class, CIFAR-10 is generally most vulnerable and MNIST is least.
        In our non-private models, we can observe an overall trend of reduced vulnerability in relation to reduced class sizes.
        From \cref{fig:class_size_f1_results} we can see that the average utility still holds up until 2000 samples before steadily declining from 1000 samples onwards.
        Here, the vulnerability advantages at 2000 samples could signify a good utility-privacy trade-off, but only from 1000 samples and lower, we can notice a major threat reduction by about half.
        With privacy ($\varepsilon=30$ and $\varepsilon=1$), vulnerability significantly decreases, making attacks less effective.
        Even our weaker privacy at $\varepsilon=30$ is able to effectively limit attack threats, while $\varepsilon=1$ models even demonstrate near-random attack effectiveness.
        Nonetheless, we find two outliers for class sizes of 50 and 100 that still show vulnerability at low TPR even in our strictly private model, which is due to the model's very low performance in these cases. The model's results are just random guessing and are therefore not really meaningful.
    
    \textbf{Class count.}
        \begin{figure}[t]
            \centering
            \begin{subfigure}[c][][c]{0.3\textwidth}
                \includegraphics[width=\textwidth]{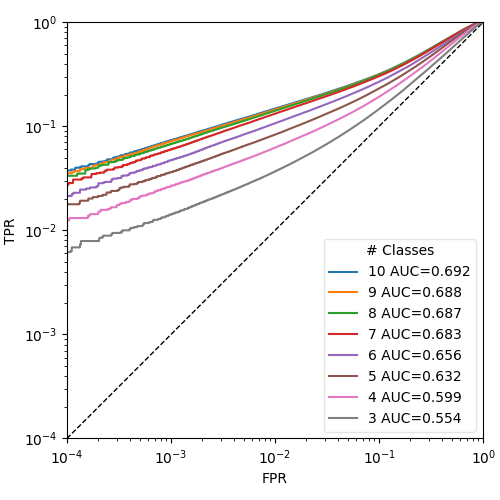}
                \caption{Non-private ($\varepsilon=\infty$) model.}
            \end{subfigure}
            \hfill
            \begin{subfigure}[c][][c]{0.3\textwidth}
                \includegraphics[width=\textwidth]{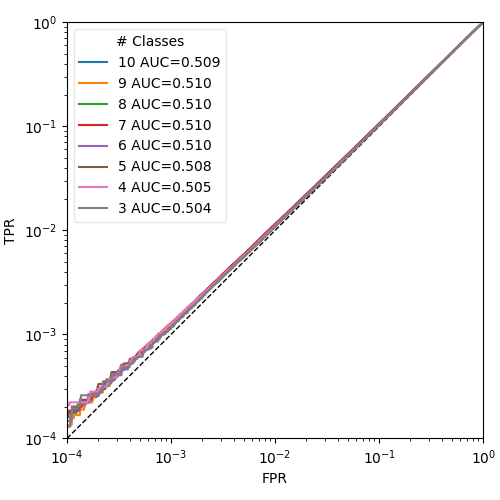}
                \caption{Private ($\varepsilon=30$) model.}
            \end{subfigure}
            \hfill
            \begin{subfigure}[c][][c]{0.3\textwidth}
                \includegraphics[width=\textwidth]{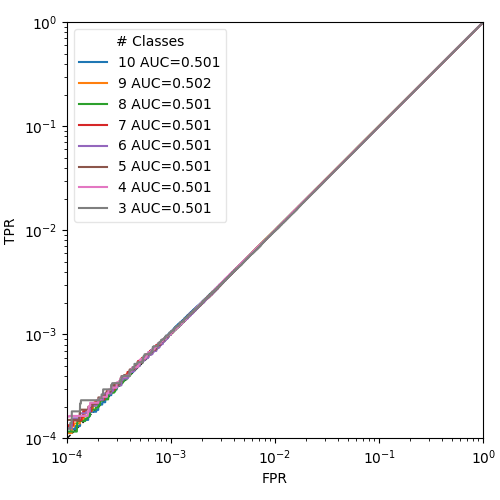}
                \caption{Private ($\varepsilon=1$) model.}
            \end{subfigure}
            \caption{
                Attack ROC curves for averaged non-private and private models with modified class counts.}
            \label{fig:class_count_attack_roc_results}
        \end{figure}
    	In this experiment, performance intuitively increases when lowering the class count, since the classification task gets easier.
    	Therefore, we find the non-private models’ average F1-score continously increasing from $0.90$ (10 classes) to $0.97$ (3 classes) and $0.68$ (47 classes) to $0.81$ (3 classes).
    	Further, the more complex tasks like CIFAR-10 and SVHN see greater benefits from reducing classes.
    	The privacy results in \Cref{fig:class_count_attack_roc_results} show that non-private models become less vulnerable as class count decreases. 
    	We just focus on reducing from 10 classes, since the 47 class case gives the same patterns. 
    	Private models generally show reduced vulnerability, with $\varepsilon=1$ models exhibiting negligible vulnerability changes because all models are very close to the optimum even at very low TPR.
        In summary, reducing class count increases utility and reduces vulnerability, with private training significantly enhancing security across all scenarios.
    
    \textbf{Class imbalance.}
        \begin{figure}[t]
            \centering
            \begin{subfigure}[c][][c]{0.328\textwidth}
                \includegraphics[width=\textwidth]{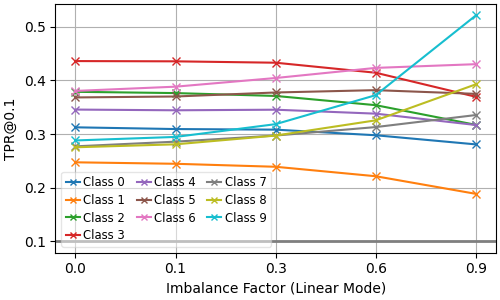}
                \caption{Non-private ($\varepsilon=\infty$) model.}
            \end{subfigure}
            \hfill
            \begin{subfigure}[c][][c]{0.328\textwidth}
                \includegraphics[width=\textwidth]{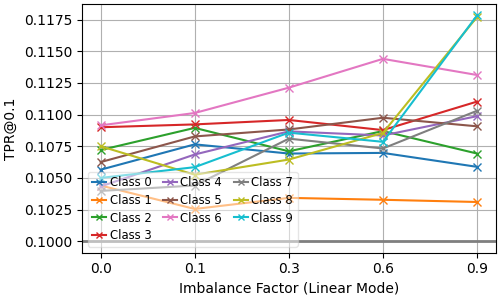}
                \caption{Private ($\varepsilon=30$) model.}
            \end{subfigure}
            \hfill
            \begin{subfigure}[c][][c]{0.328\textwidth}
                \includegraphics[width=\textwidth]{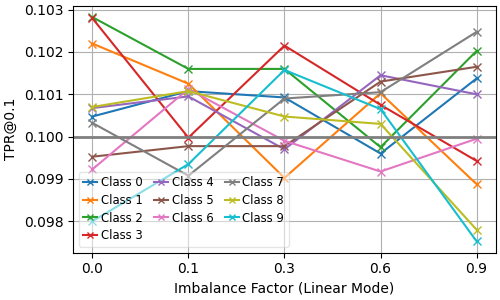}
                \caption{Private ($\varepsilon=1$) model.}
            \end{subfigure}
            \caption{
                Average class-wise attack TPR@0.1 results for datasets with varying class imbalance (linear mode) and privacy budgets.
                Note the scaling of the y-axes between different privacy budgets.}
            \label{fig:class_imbalance_fpr01_results_class_wise}
        \end{figure}
        \begin{figure}[t]
            \centering
            \begin{subfigure}[c][][c]{0.3\textwidth}
                \includegraphics[width=\textwidth]{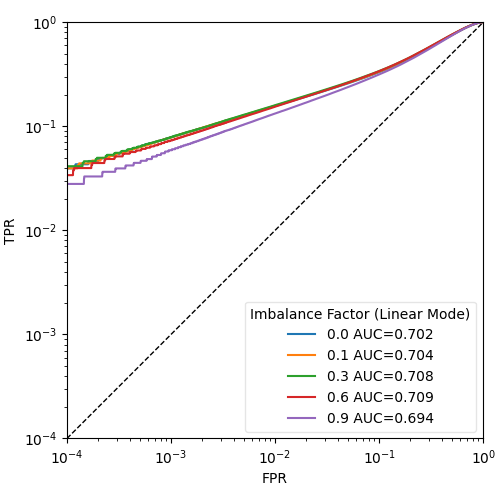}
                \caption{Non-private ($\varepsilon=\infty$) model.}
            \end{subfigure}
            \hfill
            \begin{subfigure}[c][][c]{0.3\textwidth}
                \includegraphics[width=\textwidth]{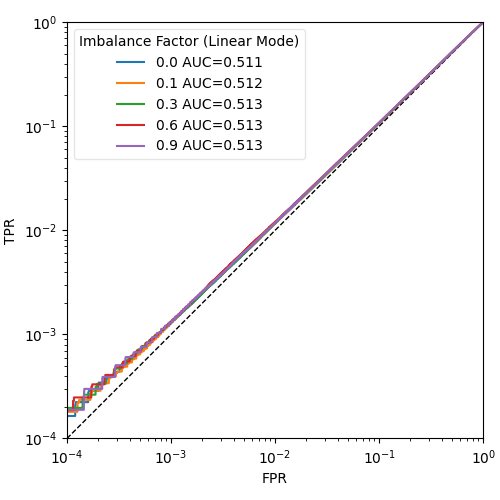}
                \caption{Private ($\varepsilon=30$) model.}
            \end{subfigure}
            \hfill
            \begin{subfigure}[c][][c]{0.3\textwidth}
                \includegraphics[width=\textwidth]{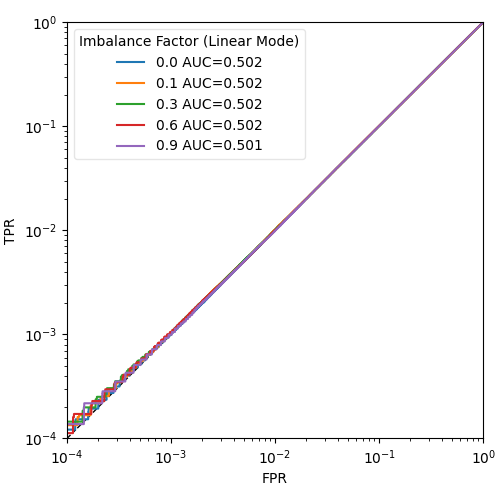}
                \caption{Private ($\varepsilon=1$) model.}
            \end{subfigure}
            \caption{
                Attack ROC curves for averaged models with modified class imbalance (linear mode).}
            \label{fig:class_imbalance_roc_curves}
        \end{figure}
        In terms of utility, increasing imbalance in linear mode linearly decreases overall F1-scores by 4\% due to the underperformance of the created minority classes, which is amplified in the private models with 8\% and 14\% at $\varepsilon=30$ and $\varepsilon=1$, respectively. Normal mode showed especially devastating utility loss (29\%, 39\%, 48\%) because the normal distribution randomness leads to having mostly minority classes and keeping just one or two bigger classes at higher imbalance factors, which shows to be very challenging. We can thus only conclude that such data setups are generally infeasible for equally weighted classification tasks, since usable performance across all classes is essential.
        
        Minority classes created further problems when looking at the linear mode attack TPR@0.1 for each class in \cref{fig:class_imbalance_fpr01_results_class_wise}, where we can clearly see the minority classes spiking in TPR for the non-private and $\varepsilon=30$ models, increasing their threat level compared to the other classes. Important to note however, that the maximum for the private model is significantly lower and therefore a less pronounced increase. For $\varepsilon=1$, the strict DP successfully obfuscates the produced minority classes, resulting in very low TPR and no recognizable outlier.
        In \cref{fig:class_imbalance_roc_curves} we can compare these results to the average ROC curves over all classes, which paints a different picture of vulnerability. When focusing on averages, we no longer see significant differences in threat levels and instead only notice slight changes in relation to shifting imbalance. We instead only record the notable changes due to DP in our privacy models.
        An increasing class imbalance thus reduces both utility and privacy due to minority classes, which is however not always clearly visible when just focusing on averages.

\subsection{Data-Level Results}
\label{subsec:Data Level Investigations Results}

\begin{table}[t]
    \small
    \begin{center}
    \begin{tabular}{lccccccccc}
        \toprule
        & & \multicolumn{2}{c}{Compression} & \multicolumn{2}{c}{$\varepsilon=\infty$} & \multicolumn{2}{c}{$\varepsilon=30$} & \multicolumn{2}{c}{$\varepsilon=1$} \\
        \cmidrule(lr){3-4}\cmidrule(lr){5-6}\cmidrule(lr){7-8}\cmidrule(lr){9-10}
        Dataset             & Entropy   & JPEG & PNG       & F1   & AUC     & F1   & AUC      & F1    & AUC   \\\midrule
        MNIST               & 0.20      & 0.52 & 0.35      & 0.99 & 0.54    & 0.98 & 0.50     & 0.95  & 0.50  \\
        FMNIST              & 0.51      & 0.57 & 0.65      & 0.93 & 0.64    & 0.89 & 0.51     & 0.85  & 0.50  \\
        SVHN                & 0.82      & 0.16 & 0.56      & 0.89 & 0.71    & 0.79 & 0.51     & 0.67  & 0.50  \\
        \dittotikz (gray)   & 0.79      & 0.33 & 0.64      & 0.88 & 0.71    & 0.79 & 0.51     & 0.66  & 0.50  \\
        CIFAR-10            & 0.89      & 0.19 & 0.73      & 0.78 & 0.86    & 0.64 & 0.52     & 0.51  & 0.50  \\
        \dittotikz (gray)   & 0.86      & 0.43 & 0.81      & 0.76 & 0.87    & 0.62 & 0.51     & 0.48  & 0.50  \\
        \bottomrule
    \end{tabular}
    \end{center}
    \caption{
        This overview presents the information density measurements together with each dataset's utility (F1-score) and vulnerability (AUC) across privacy budgets.
        A privacy budget of $\varepsilon=\infty$ indicates a non-private model and an $AUC=0.50$ translates to no vulnerability.}
    \label{tab:data_level_res}
\end{table}

We now analyse the influence of data-level properties such as information density, color, and class separability on model behavior.
Apart from information density, \cref{tab:data_level_res} also presents our datasets' overall results regarding model utility (F1-score) and vulnerability (AUC) across the three privacy levels. Regarding these DP levels, we can see the general expected trend of reduced utility with stricter budgets. This however also results in really strong defense against our attacks, where models with $\varepsilon=30$ already reduce their risks to only minor deviations from the perfect score of 0.50 AUC. Models at $\varepsilon=1$ are private enough to even guarantee perfect scores for all our datasets. 
Comparing between datasets, our models trained on MNIST exhibit the highest non-private utility at $0.99$, with a clear downwards trend as we increase the complexity of the classification task until we reach $0.78$ on CIFAR-10. Shifting to the private models with $\varepsilon=30$ and $\varepsilon=1$, we see just a slight drop to a low of $0.95$ in MNIST, while the more complex tasks see a steeper decrease in utility, like CIFAR-10 falling to $0.51$ F1.
Transitioning to vulnerability we see the same trend for non-private models, where MNIST is least vulnerable with an AUC of $0.54$ and AUC increases up to the CIFAR-10 being most vulnerable at $0.87$ AUC. However, as stated before, private learning significantly reduces vulnerability across all datasets and already almost fully removes these discrepancies at the weaker level of $\varepsilon=30$.
It will be interesting to see if we can confirm these observations using our information density, color, and class separability metrics.

\textbf{Information density.}
    We again use the results from \cref{tab:data_level_res} for evaluating our information density metrics, where we gathered each datasets entropy and JPEG/PNG compression rates. These values are accompanied by model utility (F1-score) and vulnerability (AUC) results across our three privacy levels.
	Higher shannon entropy seems to correlate with increased vulnerability. MNIST had the lowest entropy and vulnerability, while the other higher entropy datasets each increased in vulnerability up to the CIFAR-10, which showed the highest risk.
	The JPEG and PNG compression ratios do not show the same clear correlations with vulnerability or utility. Lower JPEG compression indicates a relation to higher vulnerability in SVHN and CIFAR-10, which is however undermined by their gray image variants that show increased compression ratio at the same AUC results.
	
\begin{wrapfigure}[17]{R}{0.35\textwidth}
    \vspace{-0.3cm}
    \centering
    \begin{subfigure}[c][][c]{0.35\textwidth}
        \includegraphics[width=\textwidth]{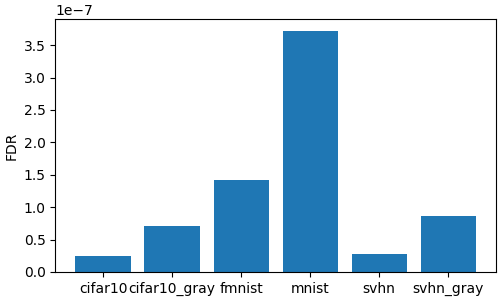}
        \caption{FDR}
    \end{subfigure}
    \begin{subfigure}[c][][c]{0.35\textwidth}
        \includegraphics[width=\textwidth]{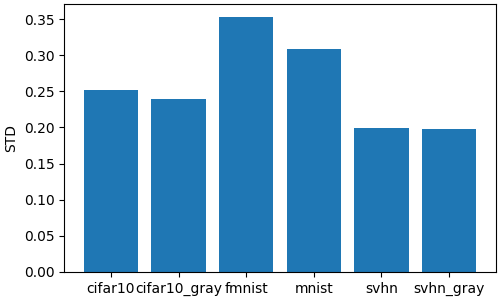}
        \caption{Average STD}
    \end{subfigure}
    \caption{
        Visualization of class separability measurement results.}
    \label{fig:class_seperability}
\end{wrapfigure} 

\textbf{Color.}
    When investigating the influence of color using our gray datasets in \cref{tab:data_level_res}, we find only slight differences in utility between the model on grayscale or color data.
    The same holds for vulnerability, where we see just a minimal difference when using grayscale, making practically no impact on our model results.
    Therefore, color does not seem to hold significant influence or potential when optimizing our data for private training.

\textbf{Class separability.}
    For analyzing class separability we still use the utility and vulnerability results from \cref{tab:data_level_res} but now try to link them to the FDR and STD results given in \cref{fig:class_seperability}.
    We clearly notice that MNIST has the highest FDR, indicating better separability between its classes.
    We find both, MNIST and FMNIST, showing higher FDR, while in turn exhibiting lower vulnerability and utility loss than the other datasets.
    On the lower end of FDR, the correlation seems to vanish, since SVHN and CIFAR-10 show to be close in FDR, while their AUC is clearly set apart by a 0.15 difference.
    In the same vein, the gray variants do not match the others regarding their FDR and AUC rankings.
    Regarding STD, we find the same results, where again the MNIST and FMNIST with low vulnerability can be successfully separated from the other datasets due to their high STD values.
    Among the other datasets we again do see the same trend, where vulnerability does not follow their STD differences.

\subsection{Discussion}\label{sec:discussion}

    \begin{table}[t]
        \caption{
            Overview of dataset-level investigation results.
            An $\varepsilon = \infty$ represents the non-private models.}
        \label{tab:dataset_level_investigation_overview}
        \begin{center}
            \small
            \renewcommand*{\arraystretch}{0.05}
            \setstretch{0.85}
            \begin{tabularx}{\textwidth}{@{}YXXX@{}}
                \toprule
                \textbf{Experiment} & $\varepsilon=\infty$ & $\varepsilon=30$ & $\varepsilon=1$ \\
                \midrule
                \textit{Class Size} ~~~~~~~~~~~~~~~~~ – Decrease Number of Samples per Classs &
                \vspace{-2mm}
                \begin{itemize}[leftmargin=0.4em, topsep=0pt, itemindent=0pt, labelsep=1pt, listparindent=0pt]
                    \setlength\itemsep{0.5em}
                    \item decreasing utility, overfitting effect appears
                    \item vulnerability increase at overfitting, else decreasing (due to low utility)
                \vspace{-2mm}\end{itemize} &
                \vspace{-2mm}
                \begin{itemize}[leftmargin=0.4em, topsep=0pt, itemindent=0pt, labelsep=1pt, listparindent=0pt]
                    \setlength\itemsep{0.5em}
                    \item utility starts decreasing earlier, smaller overfitting effect
                    \item small vulnerability increase at overfitting, else decreasing
                \vspace{-2mm}\end{itemize} &
                \vspace{-2mm}
                \begin{itemize}[leftmargin=0.4em, topsep=0pt, itemindent=0pt, labelsep=1pt, listparindent=0pt]
                    \setlength\itemsep{0.5em}
                    \item utility starts decreasing even earlier, no overfitting effect
                    \item no changes in vulnerability
                \vspace{-2mm}\end{itemize} \\\midrule
    
                \textit{Class Count} ~~~~~~~~~~~~~ – Decrease Number of Classes &
                \vspace{-2mm}
                \begin{itemize}[leftmargin=0.4em, topsep=0pt, itemindent=0pt, labelsep=1pt, listparindent=0pt]
                    \setlength\itemsep{0.5em}
                    \item utility increase, decreasing overfitting
                    \item strong vulnerability decrease
                \vspace{-2mm}\end{itemize} &
                \vspace{-2mm}
                \begin{itemize}[leftmargin=0.4em, topsep=0pt, itemindent=0pt, labelsep=1pt, listparindent=0pt]
                    \setlength\itemsep{0.5em}
                    \item more increasing utility, no decreasing overfitting
                    \item small vulnerability decrease for some datasets
                \vspace{-2mm}\end{itemize} &
                \vspace{-2mm}
                \begin{itemize}[leftmargin=0.4em, topsep=0pt, itemindent=0pt, labelsep=1pt, listparindent=0pt]
                    \setlength\itemsep{0.5em}
                    \item most increase in utility
                    \item no changes in vulnerability
                \vspace{-2mm}\end{itemize} \\\midrule
    
                \textit{Class Imbalance} – Increase Dataset Imbalance &
                \vspace{-2mm}
                \begin{itemize}[leftmargin=0.4em, topsep=0pt, itemindent=0pt, labelsep=1pt, listparindent=0pt]
                    \setlength\itemsep{0.5em}
                    \item decreasing utility (mostly minority classes)
                    \item vulnerability increase of minority classes
                \vspace{-2mm}\end{itemize} &
                \vspace{-2mm}
                \begin{itemize}[leftmargin=0.4em, topsep=0pt, itemindent=0pt, labelsep=1pt, listparindent=0pt]
                    \setlength\itemsep{0.5em}
                    \item stronger utility decrease of minority classes
                    \item smaller vulnerability increase of minority classes
                \vspace{-2mm}\end{itemize} &
                \vspace{-2mm}
                \begin{itemize}[leftmargin=0.4em, topsep=0pt, itemindent=0pt, labelsep=1pt, listparindent=0pt]
                    \setlength\itemsep{0.5em}
                    \item strongest utility decrease of minority classes
                    \item no changes in vulnerability
                \vspace{-2mm}\end{itemize} \\\bottomrule
            \end{tabularx}
        \end{center}
    \end{table}

    \begin{table}[t]
        \caption{
            Overview of the data-level investigation findings.}
        \label{tab:data_level_investigation_overview}
        \begin{center}
            \renewcommand*{\arraystretch}{1}
            \small
            \begin{tabularx}{\textwidth}{@{}p{4cm}X@{}}
                \toprule
                \textbf{Investigated Aspect} & \textbf{Observed Effects} \\ \midrule
                Entropy & higher values $\rightarrow$ more vulnerable \& less utility in private learning   \\
                JPEG Compression Ratio & higher values $\rightarrow$ less complex images \& less vulnerable \\
                PNG Compression Ratio & no recognizable effect                                              \\
                \ac{FDR} & higher values $\rightarrow$ less utility loss in private learning                \\
                \ac{STD} & lower values $\rightarrow$ higher vulnerability                                  \\
                Removal of Color & slightly worse utility                                                   \\\bottomrule
            \end{tabularx}
        \end{center}
    \end{table}

    We first want to summarize our overall findings before giving our extracted practical guidelines.
    Starting with the class size experiment in \cref{tab:dataset_level_investigation_overview}, reducing the samples per class decreases utility for both private and non-private models, with private models being more sensitive. Overfitting is observed in non-private models and slightly in private models with $\varepsilon=30$. The most private model ($\varepsilon=1$) showed no overfitting. Vulnerability initially increases with fewer samples and then decreases, linked to the overfitting effect.
    Reducing the number of classes increases model utility across all models, with a more pronounced effect in models with lower privacy budgets. Higher class counts tend to make models more vulnerable, whereas fewer classes reduce this vulnerability, especially in non-private models. Although it might seem counterintuitive---since more classes imply more data to obscure sensitive information---it is in line with \citet{Shokri_Stronati_Song_Shmatikov_2017} and the key insight, that models are generally less confident when classifying a larger number of classes. This reduced confidence makes it easier to distinguish between seen and unseen samples, where the confidence usually spikes in previously seen ones. Conversely, with fewer classes, models exhibit higher overall confidence due to the simpler classification task, reducing the confidence gap between seen and unseen, and thus lowering the attack's success.
    Finally, for class imbalance, we notice that model utility decreases with increasing imbalance, especially for private models. Minority classes are more vulnerable, with this effect being smaller in private models with $\varepsilon=30$ and not observed in models with $\varepsilon=1$.
    
    For the data-level results presented in \cref{tab:data_level_investigation_overview}, we first look at the entropy.
    The datasets with higher entropy are more vulnerable and lose more utility with private learning.
    However, we can only find these results in the average entropy over the entire datasets, whereas a class-wise comparison between the entropy and attack proneness of individual classes does not show any correlation.
    The JPEG compression ratio can instead help to estimate the complexity of an image dataset and by that also its task difficulty.
    This is supported with higher ratios indicating higher utility and lower vulnerability in our models.
    In terms of class separability, we find that the \ac{FDR} can provide a general indication of model vulnerability.
    Datasets with high \ac{FDR} are less vulnerable and lose less utility with private learning.
    The average dataset \ac{STD} is also related to vulnerability, with lower \ac{STD} datasets being more likely to be vulnerable.
    Converting colored datasets to grayscale slightly reduces utility and marginally influences vulnerability. This can be linked to the decreased entropy, altered JPEG ratio, and \ac{FDR} values, though these metrics do not consistently predict the vulnerability changes.
    Especially the color influence experiments show that the data-level metrics and their effect on vulnerability and utility can only be used as general indicators.
    
    \subsubsection{Implications}\label{sec:imp}
        No single data-related characteristic fully describes how a dataset affects a private model’s performance or vulnerability.
        Instead, these metrics must be combined to estimate the model’s behavior when trained on a specific dataset.
        We untangled five key rules of thumb regarding dataset characteristics for building private machine learning applications:
    	(1)	Use DP-private learning: In sensitive contexts, apply \ac{DP}-private learning whenever possible. We find that even with a larger privacy budget, attack success is significantly reduced, balancing vulnerability differences between models trained on different datasets.
    	(2)	Amount of data: Fewer samples per class also reduce the overall data, leading to overfitting and increased vulnerability. Prioritize acquiring data where possible, especially for smaller classes, over optimizing the training process.
    	(3)	Number of classes: More classes increase the proneness to \ac{MIA}s. Reduce the number of classes to a feasible minimum for the task or use \ac{DP}-private learning to mitigate this vulnerability.
    	(4)	Dataset imbalance: Imbalance affects individual class vulnerability, especially for minority classes. Balance the dataset manually or use private models to equalize vulnerability across classes.
    	(5)	Image complexity: More complex datasets have a worse utility-privacy trade-off and higher vulnerability. Analyzing entropy, \ac{FDR}, \ac{STD}, and JPEG compression ratio can help estimate dataset complexity. These metrics are useful for comparing datasets and assessing how models change with new data.

	\begin{table}[t]
        \small
        \begin{center}
        \begin{tabular}{cccccccccccc}
            \toprule
            & & \multicolumn{2}{c}{Compression} & \multicolumn{2}{c}{Separability} & \multicolumn{2}{c}{$\varepsilon=\infty$} & \multicolumn{2}{c}{$\varepsilon=30$} & \multicolumn{2}{c}{$\varepsilon=1$} \\
            \cmidrule(lr){3-4}\cmidrule(lr){5-6}\cmidrule(lr){7-8}\cmidrule(lr){9-10}\cmidrule(lr){11-12}
            Size   & Entr.   & JPEG & PNG     & FDR     & STD     & F1   & AUC     & F1   & AUC      & F1    & AUC   \\\midrule
            7.2k   & 0.94      & 0.03 & 0.19    & $<$0.01 & 0.25    & 0.92 & 0.64    & 0.85 & 0.53     & 0.79  & 0.51  \\
            \bottomrule
        \end{tabular}
        \end{center}
        \caption{
            Results for the practical privacy scenario on the more sensitive COVID-19 data. We look at some characteristics, utility (F1-score), and vulnerability (AUC) across privacy budgets.
            A privacy budget of $\varepsilon=\infty$ indicates a non-private model and an $AUC=0.50$ translates to no vulnerability.}
        \label{tab:covid}
    \end{table}
	
	\subsubsection{Comparison to Practical Privacy Scenario}\label{sec:practice}
    	
        Before concluding, we compare our findings to a practical test scenario.
        A limitation of our analysis is that it was conducted solely on benchmarking datasets, which, while fitting for our study, lack obvious privacy implications (\cref{sec:Datasets}).
        To address this, we test our results in a privacy-conscious setting by drawing on previous work~\cite{langePrivacyPracticePrivate2023}, where we investigated private COVID-19 detection.
        Unlike our benchmarking datasets, this binary task involved sensitive medical images, where a successful \ac{MIA} could reveal a person’s COVID-19 status.
        
        Considering our rules (1)--(5) from \cref{sec:imp} and the characteristics of the COVID-19 dataset~\cite{chowdhury2020can,rahman2021exploring}—(2) limited training data, (3) two classes, (4) medium imbalance (50\% more normal than COVID-19 cases), and (5) high complexity with low compression and FDR (see \cref{tab:covid})—we expect a challenging utility-privacy trade-off and vulnerability to \ac{MIA}s, though the low class count (3) might provide some relief.
        The results in \cref{tab:covid} confirm medium vulnerability at $\varepsilon=\infty$ and a noticeable utility drop with stricter DP budgets on the COVID-19 model.
        Despite more challenging data-level statistics compared to the FMNIST dataset (see \cref{tab:data_level_res}), both show roughly equal results, likely due to the 2- vs. 10-class task advantage.
        Using our suggestions, although we cannot collect more data (2)+(4) or further reduce classes (3), we can apply strategy (1) and successfully use a lower \ac{DP} level of $\varepsilon = 30$ at a low $AUC=0.53$, maintaining both utility and defense capabilities.

\section{Conclusion}
\label{sec:conclusion}

In this work, we take into account various dataset characteristics to provide guidance for implementing differentially-private image classification models.
The primary goal is to help researchers and practitioners in determining a priori, if \ac{DP} is worthwhile and what needs to be considered for specific datasets.
In our experiments, we assess \ac{ML} model utility and vulnerability to \ac{MIA}s across different datasets, while relying on varying privacy budgets for \ac{DP} ($\varepsilon = \{\infty, 30, 1\}$).
Our derived implications for effectively using our results in engineering private \ac{ML} models are summarized in \cref{sec:imp} and our findings have been applied to a practical scenario in \cref{sec:practice}.
We want to give data-related optimization a bigger stage in \ac{PPML}, where we might not be able to directly access private data but instead have to rely on general data metrics.
A key aspect of our analysis is that our models effectively mitigate most of the MIA threat across all datasets with a modest privacy budget of $\varepsilon = 30$, achieving a more practical utility-privacy trade-off at low risk.
By considering the influence of different aspects of a dataset in addition to the actual training process, the broader picture allows steering in the most effective directions.
%
%
Future work could explore the influence of model architectures on \ac{MIA} threat and extend this study to non-image data.

\textbf{Acknowledgments.}
The authors acknowledge the financial support by the Federal Ministry of Education and Research of Germany and by the Sächsische Staatsministerium für Wissenschaft, Kultur und Tourismus for ScaDS.AI.
Computations for this work were done (in part) using resources of the Leipzig University Computing Centre.

\printbibliography





\end{document}